\documentclass[letterpaper, 10 pt, conference]{ieeeconf}  

\IEEEoverridecommandlockouts                              

\newcommand{\symbolRotationAdjustment}{\alpha_{corr}}
\newcommand{\symbolTranslationAdjustment}{y_{corr}}

\usepackage{graphicx} 
\usepackage{amsmath} 
\usepackage{amssymb}  
\usepackage{array}
\usepackage{booktabs} 
\usepackage{soul}
\usepackage{svg}
\usepackage{color}

\usepackage[ruled, linesnumbered]{algorithm2e}

\SetKwProg{Fn}{Function}{}{end}\SetKwFunction{FRecurs}{FnRecursive}%
\SetAlgoLongEnd

\title{\LARGE \bf
RSV: Robotic Sonography for Thyroid Volumetry
}

\author{John Zielke$^{*,1}$, Christine Eilers$^{*,1}$, Benjamin Busam$^{1}$, Wolfgang Weber$^{3}$, \\ 
Nassir Navab$^{1},^{2}$ and Thomas Wendler$^{1}$
\thanks{*Both authors contributed equally.}
\thanks{**This work was partially supported by the EU grant 688279 (EDEN2020).}
\thanks{$^{1}$J. Zielke, C. Eilers, B. Busam, N. Navab and T. Wendler are with the Chair for Computer Aided Medical Procedures \& Augmented Reality,
        Technical University of Munich, 85748 Garching near Munich, Germany
        (email: john.zielke@tum.de, christine.eilers@tum.de, b.busam@tum.de, nasssir.navab@tum.de, wendler@tum.de)}%
\thanks{$^{2}$N. Navab is with the Whiting School of Engineering, John Hopkins University, Baltimore, MD, USA (email: nassir.navab@jhu.edu)}%
\thanks{$^{3}$W. Weber is with the Department of Nuclear Medicine, Technical University of Munich, Munich, Germany (email: w.weber@tum.de)}%
}

\begin{document}

\maketitle
\thispagestyle{empty}
\pagestyle{empty}

\begin{abstract}

In nuclear medicine, radioiodine therapy is prescribed to treat diseases like hyperthyroidism. The calculation of the prescribed dose depends, amongst other factors, on the thyroid volume. This is currently estimated using conventional 2D ultrasound imaging. However, this modality is inherently user-dependant, resulting in high variability in volume estimations. To increase reproducibility and consistency, we uniquely combine a neural network-based segmentation with an automatic robotic ultrasound scanning for thyroid volumetry. The robotic acquisition is achieved by using a 6 DOF robotic arm with an attached ultrasound probe. Its movement is based on an online segmentation of each thyroid lobe and the appearance of the US image. During post-processing, the US images are segmented to obtain a volume estimation. In an ablation study, we demonstrated the superiority of the motion guidance algorithms for the robot arm movement compared to a naive linear motion, executed by the robot in terms of volumetric accuracy. In a user study on a phantom, we compared conventional 2D ultrasound measurements with our robotic system. The mean volume measurement error of ultrasound expert users could be significantly decreased from $20.85\pm16.10\%$ to only $8.23\pm3.10\%$ compared to the ground truth. This tendency was observed even more in non-expert users where the mean error improvement with the robotic system was measured to be as high as $85\%$ which clearly shows the advantages of the robotic support. \\

\textbf{\textit{Index Terms} - Robotic ultrasound, thyroid volumetry, online segmentation}

\end{abstract}

\section{Introduction \& state of the art}

\begin{figure}
   \centering
   \includegraphics[width=\linewidth]{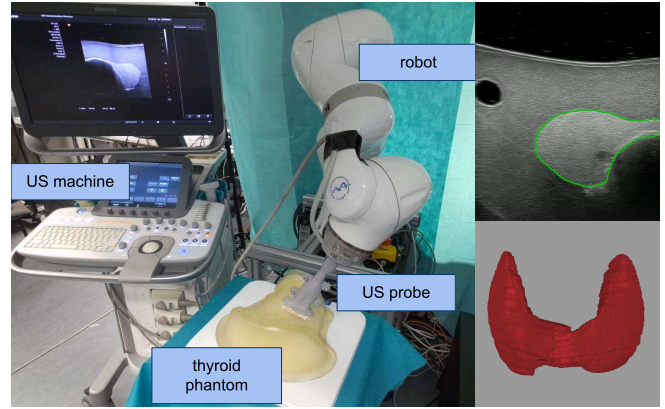}
    \caption{An overview of the proposed robotic system. Left: the hardware setup with an US machine, a 6 DoF robotic arm, an US probe attached to the robot end-effector and a thyroid phantom, top right: the online 2D segmentation of the thyroid during a robotic scan, bottom right: the final 3D thyroid segmentation for both lobes.}
    \label{fig:header}
\end{figure}

Hyperthyroidism is prevalent in up to 2.5\% of the population in iodine sufficient and up to 2.9\% in iodine-deficient countries~\cite{Taylor}. Patients with this disease show an overproduction of thyroid hormones, resulting in, amongst others, sweating, nervousness, weight loss, and increased and irregular heartbeats. To reduce the hormone production, radioiodine therapy (RIT) is prescribed. Here, a small dose of radioactive iodine, in most cases \textsuperscript{131}I, is administered to the body orally, concentrating in the thyroid gland and decimating thyroid cells. The dosimetry for this therapy is based on Marinelli's formula Ò(1)Ó, which takes into account the thyroid mass ($m$ in $g$), the target dose of \textsuperscript{131}I to the thyroid ($D$ in $Gy$) as well as its 24-hour uptake ($IU_{24h}$, unit-less) and its effective half-life time in the thyroid gland ($T_{eff}$ in $h$)~\cite{Szumowski}.

\begin{align}
    A = \dfrac{25 * m * D }{IU_{24h} * T_{eff}} 
\end{align}

\noindent where $25$ is a unit conversion coefficient and $A$ is the administered activity in $GBq$. The thyroid mass is derived from the thyroid volume assuming a constant density. An over- or underestimation of this volume could therefore lead to an in- or over-effective RIT.\\
Currently, volumetry estimations are obtained with conventional two-dimensional ultrasound (2D US) imaging and an ellipsoid formula approximation~\cite{Brunn}. However, 2D US is difficult to interpret and inherently user-dependant~\cite{Kojcev}. On the other hand, including a robotic collaborator can increase accuracy and repeatability~\cite{Kojcev,esposito2016multimodal}. \\
Some robotic US work has been accomplished with regards to the thyroid as organ of interest. Kim et al.~\cite{kim} developed a control algorithm based on position and force feedback, for an US scanning robot. The pressure force applied to the thyroid phantom was controlled by analyzing the root mean square error (RMSE) between consecutive images. Then a support vector machine (SVM) was trained to find optimal force US images based on the RMSE input. The robot followed a predefined trajectory and the final evaluation of the image quality was performed through rating by a clinician. Huang et al.~\cite{huang} introduced an automatic US scanning for 3D imaging with a robotic arm. Two force sensors applied to the probe enabled a force feedback control loop while the path was defined manually. The system was tested on a thyroid phantom, amongst others. Kaminski et al.~\cite{kaminski} performed a feasibility study regarding robot-assisted US scanning for the assessment of thyroid diseases. The robot was controlled with force feedback and the trajectory was planned based on a registered anatomy model. The scan repeatability was measured using the cross-correlation of images between different runs. Lastly, Kojcev et al.~\cite{Kojcev} compared the thyroid length measurements between a robotic acquisition and a manual expert measurement. The robot was controlled with force feedback. A 3D camera was used to capture a point cloud of the patient and a region of interest (ROI) was manually selected on this point cloud. A trajectory line was then computed based on the ROI before the scan. Results showed more repeatable measurements for the robotic approach compared to the manual procedure. \\ 
Even though these works show successful robotic US scans of the thyroid, in all cases, the trajectory is defined with significant human interaction, is linear and is independent of the acquired US images. The acquired data (US sweep and robot position) has not been used online to define or correct the trajectory for the particular application, i.e., the 3D reconstruction of the thyroid. Moreover, the existing literature fails to derive clinically relevant information out of the 3D reconstructed image. \\
In contrast to the state of the art, in our setup, a 6 DoF robotic arm, equipped with an US probe automatically scans the thyroid lobes based on an online segmentation. Motion guidance algorithms are applied using the acquired US image and the pose of the robot end-effector. Afterwards, the US sweep is compounded and the thyroid is automatically segmented on each B-mode image by a deep neural network. As a result, the volume of the thyroid gland is computed. To the best of our knowledge, this is the first fully implemented  pipeline on robotic US scanning for thyroid volumetry. The main contributions of this works can be summarized as follows: 
\begin{itemize}
    \item We present a full and automatic pipeline to accurately compute a 3D US image of the thyroid and its volume based on a robotic scan. 
    \item We propose image-guided methods to define the scanning trajectory, i.e., the movement of the US probe attached to the robot end-effector, based on the online acquired US images. We further include an ablation study to evaluate their impact on the calculated volume.
    \item We perform an extensive evaluation on a thyroid phantom with experts and non-experts to compare the proposed robotic method to conventional 2D US in terms of volumetric accuracy.
\end{itemize}

\section{Methodology}
This section describes the proposed methods for the robotic US scans and the thyroid volumetry. The overall workflow is described in Section~\ref{subsec::roboticSystem}, followed by a presentation of the robotic motion planning and execution in Section~\ref{subsec::Movement}. Section~\ref{subsec::volumetry} describes the automatic volumetry estimations based on a deep neural network.

\subsection{Robotic US System}
\label{subsec::roboticSystem}
\begin{figure}
   \centering
   \includegraphics[width=\linewidth]{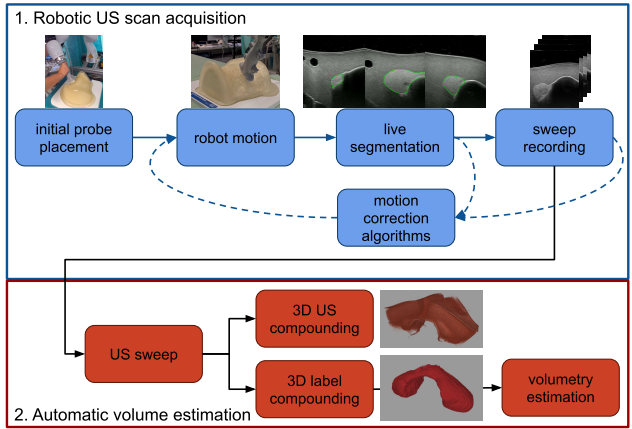}
    \caption{Our proposed workflow consists of two steps: a robotic US scan acquisition and a post-processing stage for the automatic volume estimation. The beginning of the first step involves human interaction: the user places the US probe on each lobe such that the thyroid is visible in the US image. The robot then moves the US probe while analysing the live US image. If shadows are detected or the thyroid is not centered the robot will modify the trajectory. On each step the system also evaluates if thyroid tissue is segmented. Once the end of the thyroid is reached the robot reverses the direction to acquire a sweep with the complete thyroid lobe. Finally, both the US B-mode images and the labelled US sweep are compounded (post-processing). These steps are repeated for each lobe. The merged label compounding for both thyroid lobes is used to estimate the total volume. The compounded 3D US image can be stored for potential follow-up examinations.}
    \label{fig:workflow}
\end{figure}

To address the challenge of accurate and repeatable thyroid volume estimations, we propose an automatic robotic US system pipeline (see Figure~\ref{fig:header}). The workflow is depicted in Figure~\ref{fig:workflow}. \\
\textit{1) Hardware:} The system consists of a robotic manipulator (LBR iiwa 14 R830, KUKA GmbH, Augsburg, Germany) and a Siemens Acuson NX3 US machine (Siemens Healthineers GmbH, Erlangen, Germany). A VF12-4 $12$~MHz linear probe is attached to the end-effector of the robotic arm. B-mode images ($30$~fps) are forwarded via a framegrabber connected to the control computer by a USB interface. The robot is controlled with ROS through iiwa stack~\cite{hennersperger2017towards}\footnote{\texttt{https://github.com/IFL-CAMP/iiwa\_stack}}. Visualization and control of the framework is implemented as a plugin in ImFusion Suite Version 2.13.9 (ImFusion GmbH, Munich, Germany). \\
\textit{2) Workflow:} The workflow consists of two main parts: 1) a Robotic US scan acquisition, 2) an automatic volume estimation. First, to enable the robotic scan, the US probe has to be placed on the neck by a human operator with the thyroid lobe visible on the US image. The robot then automatically moves to one end of the thyroid lobe based on the online segmentation obtained using a QuickNAT~\cite{guha_roy_quicknat_2019} network trained to segment the thyroid in the phantom. Once the US probe reaches the first lobe end, the data recording for US images and US probe positions begins and the robot motion is reversed, moving to the other end of the lobe. During the scan, the system corrects its planned path based on the live US image and the corresponding segmentation to ensure consistent image quality and a reliable acquisition of the whole thyroid volume (see \ref{subsec::Movement}).
After executing this scan procedure on both lobes with the patient assumed to be stationary, the individual scans of both lobes can be combined into one volume that is then ready for further analysis, such as volumetry or follow-up evaluations.

\subsection{Robotic Motion Planning}
\label{subsec::Movement}

\begin{figure}
   \centering
   \includegraphics[width=\linewidth]{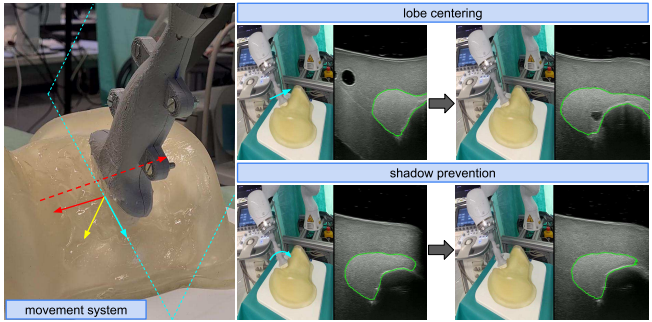}
    \caption{Overview of the robotic movement. Left: the initial coordinate system is derived from the initial US probe position. The vector normal to the US plane (red, x-axis) gives the movement direction of the US probe. The robot then moves in x-direction and after detecting the first lobe end it moves in negative x-direction (red dotted arrow). Constant force is applied through an impedance control in z-axis (yellow) and the motion corrections are performed in the y-z plane (cyan dotted). Top right: path adjustment for lobe centering by translating the probe in the y-z plane (before $\rightarrow$ after), bottom right: path adjustment for shadow prevention by rotating the probe in the y-z plane (before $\rightarrow$ after).}
    \label{fig:robotmovement}
\end{figure}

\subsubsection{Basic Movement}
The robot is controlled with a built-in, real-time impedance control (KUKA Sunrise.OS 1.15.0.9). This allows for a constant force onto the surface and removing the probe from the phantom easily. On top of this impedance control, we added the image-based motion adjustments, namely shadow prevention and lobe centering, described the in the next subsubsection. 
After the initial manual placement of the probe on the neck, the initial scan path of the robotic arm is planned along a line perpendicular to the probe pose (see Figure~\ref{fig:robotmovement}).  The robot then moves in step sizes of 0.5cm along this axis (x-axis, see Figure~\ref{fig:robotmovement} red axis). After each step, the live segmentation is analyzed with regards to the presence of thyroid in the current US image. The system considers the thyroid to be present if the segmentation of all images received in the last second contain thyroid. If this is not the case one end of the thyroid has been reached. After reaching an end of the thyroid for the first time, the movement direction is reversed and the recording of the end-effector poses and B-mode images starts. Once no presence of thyroid is detected for a second time, it is assumed that the other end of the thyroid lobe has been reached. The recording is stopped and the acquired data is post-processed to compute the lobe volume (see~\ref{subsec::volumetry}). This procedure results in scanning each thyroid lobe partially twice, depending on the initial probe placement. However, the data recording for the volume compounding is only executed in between the detection of both lobe ends, i.e. in the full continuous sweep from top to bottom.

\subsubsection{Force Feedback and Path Adjustments}
During the scan, the robot is set to move freely in x-direction while maintaining a constant force of the probe onto the phantom (z-direction). Additionally, the pose is corrected after each movement step using: i) a rotation around the principal movement axis (x-axis) to minimize shadows in the US images caused by partial contact with the probe (see Figure~\ref{fig:shadow_non_center} c) and d)), and ii) a sideways correction (y-z plane) to achieve maximum coverage of the thyroid on the B-mode image. The latter prevents partial coverage of the thyroid if the probe is not positioned in the image center, or the planned path is not aligning well with the thyroid position inside the neck (see Figure~\ref{fig:shadow_non_center} a) and b)).
Both adjustments are executed iteratively - following the algorithm below - until a satisfactory position is set or a maximum adjustment angle or offset is reached.  

\begin{algorithm}[]
 \SetKwInOut{Input}{in}
 \Input{$\symbolRotationAdjustment$, $\alpha_{corr\_max}$, $\symbolTranslationAdjustment$, $y_{corr\_max}$}
 \SetKwProg{Fn}{Function}{ begin}{end}
 \Fn{doPoseAdjustment()}{
  \While{rotationAdjustmentNeeded()}{
   \eIf{ getShadowPosition() == left}{
    rotationAngle = $\alpha_{corr\_step}$\;
   }{
    rotationAngle = -$\alpha_{corr\_step}$\;
    }
   $\alpha_{corr\_new}$ = min(max($\symbolRotationAdjustment$  + rotationAngle,   $-\alpha_{corr\_max}$),$+\alpha_{corr\_max}$)\;
   \If{$\alpha_{corr\_new}$ == $\symbolRotationAdjustment$}{
    }
    
   $\symbolRotationAdjustment$ = $\alpha_{corr\_new}$\;
   rotateAroundPrincipalAxis($\symbolRotationAdjustment$)\;
  }
  \While{ positionAdjustmentNeeded()}{
    $y_{corr\_step}$ = getYTranslationAdjustment()\;
    $y_{corr\_new}$ = min(max($\symbolTranslationAdjustment$ + $y_{corr\_step}$,
    $-y_{corr\_max}$),$y_{corr\_max}$)\;
    \If{$y_{corr\_new}$ == $\symbolTranslationAdjustment$}{

    }
     
    $\symbolTranslationAdjustment$ = $y_{corr\_new}$\;
    moveInYDirection($\symbolTranslationAdjustment$)\;
    }
    }
    \caption{Adjustment algorithm}
\end{algorithm}

\begin{algorithm}[]
 $P_{initial}$ = getCurrentPose()\;
 \Fn{ moveUntilEnd (direction)}{
  \While{checkThyroidInCurrentImages()}{
   moveStepAlongPrincipalAxis($P_{initial}$,
   stepSize * direction)\;
   doPoseAdjustment()\;
   }
  }
 moveUntilEnd(-1)\;
 startRecording()\;
  \tcc{move probe into thyroid again}
 moveStepAlongPrincipalAxis($P_{initial}$,
   stepSize)\;
 moveUntilEnd(1)\;
 stopRecording()\;
 startPostProcessing()\;
 \caption{Scan algorithm}
\end{algorithm}

\paragraph{Shadow prevention}
To determine the necessity of a rotation adjustment, the US images are analysed with an intensity-based method. A margin of the left and right side of the image are each divided into equal parts. If a certain percentage of pixel intensities in all segments on one side are below a certain threshold value ($p_{brightness}$) the probe is rotated step-wise in that direction by $\alpha_{corr,step}$. The total correction rotation at the current step is then referred to as $\symbolRotationAdjustment$ (see Algorithm 1). If both sides have insufficient contact, no rotation adjustment is performed, and a warning is send to the user.

\paragraph{Lobe Positioning in the US Image}
To determine whether an adjustment in the y-z plane of the probe is needed (see Figure~\ref{fig:robotmovement}), the live segmentation is analyzed. This live segmentation is seen as a binary label map, consisting of thyroid label (1) and background (0). The left and right side of the label map are checked for the thyroid label. If only one side contains a thyroid label, the image is adjusted by translating the probe sideways, such that the outer edge of the thyroid label is close to the respective vertical image border  (see Algorithm 1). If, for example, the right lobe is scanned and no thyroid label is detected in the left margin of the label map, the US probe will move incrementally in negative y direction in the y-z plane (Fig.~\ref{fig:robotmovement}). A defined margin of the width of the image remains on each side. This prevents an excessive motion correction based on minor changes in the image. The procedure ensures that the lobe is scanned entirely, including the biggest possible area of the isthmus (i.e., the tissue bridge connecting both thyroid lobes). 
If both edges of the segmentation contain thyroid tissue, the center of mass of the segmented thyroid within the image is calculated. The probe is then translated to move the center of mass to the center of the image. This will then move the probe to cover more of the lobe and less of the isthmus as the lobe is assumed to be bigger, eventually leading to the case where only one side of the image contains thyroid.  While it would also be possible to either translate to the right or the left edge of the thyroid based on the lobe that is scanned, analogous to the case of only one edge containing thyroid tissue, this method also ensures that in case the thyroid is wider than the coverage of the probe as much thyroid area as possible can be captured. It also does not rely on the external information of which side of the thyroid is currently scanned. The total amount of correction in y-direction of the probe at the current step is then referred to as $\symbolTranslationAdjustment$.
The desired pose of the robot during the scan can then be determined using the following formula:
\begin{align}
\textbf{R}_{corr} &= \begin{bmatrix}
1 & 0 & 0\\
0 & \cos{(\symbolRotationAdjustment)} & -\sin{(\symbolRotationAdjustment)} \\
0 & \sin{(\symbolRotationAdjustment)} & \cos{(\symbolRotationAdjustment)}
\end{bmatrix}\\
\textbf{R}_{probe} &= \textbf{R}_{corr} \cdot \textbf{R}_{init}\\
\text{t}_{probe} &= \text{t}_{init} + \textbf{R}_{init} [s \cdot n_{steps},\symbolTranslationAdjustment,0]^\text{T},
\end{align}

\noindent where $\textbf{R}_{corr}$ is the rotation matrix defined by the current rotation adjustment angle,
 $\textbf{R}_{init}$ is the probe orientation after the initial manual placement on the phantom,
 $\text{t}_{init}$ is the probe position after the initial manual placement on the phantom, 
 $n_{steps}$ is the current step position from the initial position along the scan path,
 $\text{t}_{probe}$ and $\textbf{R}_{probe}$ represent the target position and orientation of the probe at the current time step.
 
All motion steps with movement corrections are executed in real-time as long as the system does not detect the end of the scanned lobe. Once that point is reached, the end-effector is moved away from the phantom in z-direction to relieve the pressure of the probe on it. The same procedure is then repeated on the other side of the neck.

\subsection{Volumetry Estimations}
\label{subsec::volumetry}
Once the scan has been acquired, the data is post-processed. The acquired data consists of the actual US probe poses from the scanning movement and the acquired US images. This includes a 3D US compounding, a 3D label compounding and a thyroid volume estimation. To compound the volume, all recorded 2D US images are combined with the associated probe poses using their respective timestamps. As the US images and probe poses are not sampled regularly in time, the probe poses are interpolated to match the B-mode image timestamps. A linear interpolation is used for the translation of the probe, while a spherical linear interpolation is used for the probe rotation~\cite{busam2016quaternionic}. These pose and image pairs are then sent to a compounding algorithm. The same steps are also applied on the thyroid masks generated during the sweep which results in a 3D segmentation mask $M_{thyroid}$. After the scan procedure has been executed on both lobes, the two individual volumes that are returned by the compounding algorithm are merged through a spatially correct union of both lobe segmentation. The fused 3D label map is finally used to calculate the total thyroid volume. In this combined volume, it holds
\begin{align}
M_{thyroid}(i,j,k) = \begin{cases}
1 & v_{i,j,k}\ \text{contains thyroid}\\
0 & \, \text{otherwise.}
\end{cases}
\end{align}
for each voxel $v_{i,j,k}$, where $i, j$ and $k$ are the dimensions of the label map.
The total volume of the thyroid within a 3D volume $v$ of $l \cdot w \cdot h$ voxels with voxel dimension $d_1,d_2,d_3$ is then given by:
\begin{align}
V_{thyroid} = d_1 \cdot d_2  \cdot d_3 \cdot \sum_{i=1}^{l}\sum_{j=1}^{w}\sum_{k=1}^{h} M_{thyroid}(i,j,k)
\end{align} 
The values for each voxel dimension ($d_1,d_2,d_3$) are returned by the compounding algorithm.

\section{Experiments and Results}
\subsection{Experimental Setup}
The experimental setup reflects the general setup of the robotic US system (see Fig.~\ref{fig:header}, left). The evaluation was performed on a thyroid phantom (thyroid ultrasound training phantom, model 074, CIRS, USA). For the shadow prevention both image sides (with a margin of 5 \% of the total image width) are divided into 8 segments each. If $90 \%$ of the pixel values in a segment lie below $p_{brightness} = 70$ (value range $0 - 255$), the segment is defined as containing a shadow. If a shadow is detected, rotations are applied in steps of $\alpha_{corr\_step} = 5^{\circ}$ until a maximum of $\alpha_{corr\_max} = 30^{\circ}$. For the lobe centering, we analyze the borders of the image with a size of $4 \%$ of the total image width. If not centered, the transducer will be moved in a way such that the outer lobe contour is located $6 \%$ of the total image width away from the image border. The maximum offset is set to $y_{corr,max} = 8 cm$.

Additionally, a second hand-held US-probe (VF12-4 linear probe) is provided for conventional 2D US scanning. 12 volunteers participated in the evaluation, who can be divided into 5 experts and 7 non-experts. Ground truth for all evaluations is a CT scan of the phantom (voxel size $0.625 \times 0.625 \times 0.8 mm^3$) with a manually segmented thyroid volume of $30.01 ml$.

\subsection{Influence of motion corrections}
To analyze the influence of the proposed motion correction methods, we performed an ablation study. For this, 74 initial probe positions (37 scans with two lobes each) were recorded. We then performed US sweeps on the phantom in four different configurations: no motion correction, only shadow prevention, only centering and both corrections. This amounted in a total of 296 scans and 148 complete thyroid volume estimations. All scans and segmentation masks were compounded and the volume was calculated per thyroid. Quantitatively, an improvement in the volume calculation can be seen between no correction and the shadow prevention as well as between only shadow prevention adjustments and both adjustments combined. The translation adjustment alone performed worse than no motion correction. One reason for this could be an introduction of shadow artifacts. By centering the thyroid lobe in the image, the US probe can loose contact to the phantom surface and in this way worsen the scan quality (Table~\ref{table_motion_correction}). Qualitative results are shown in Figure 5.

\begin{figure}
   \centering
   \includegraphics[width=\linewidth]{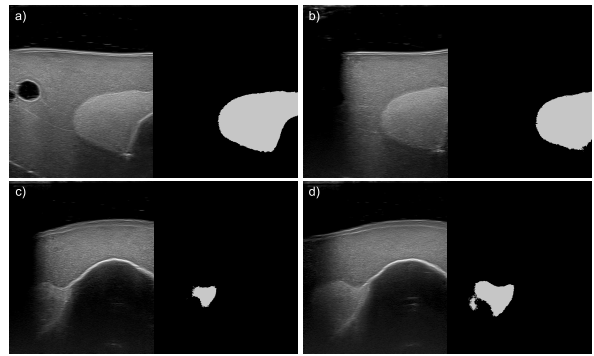}
    \caption{Examples of suboptimal US scans (left) with their online estimated segmentation masks (right). Top row, a) and b): the lobe is not centered, bottom row, c) and d): shadowing prevents optimal segmentation results.}
    \label{fig:shadow_non_center}
\end{figure}

\begin{table}[h]
\caption{The influence of motion corrections on the volumetry estimations. The initial positions of 37 scans were used to acquire scans in all four correction configurations.}
\label{table_motion_correction}
\begin{center}
\begin{tabular}{rcc}
\toprule
type of motion correction & average volume (ml) & $\pm$std (ml) \\
\midrule
none & 29.30 & 3.63 \\
only shadow prevention & \textbf{30.97} & 1.69\\
only centering & 29.05 & 4.60 \\
shadow prevention \& centering & 31.90 & \textbf{1.00} \\
\bottomrule
\end{tabular}
\end{center}
\end{table}

\begin{figure}
   \centering
   \includegraphics[width=\linewidth]{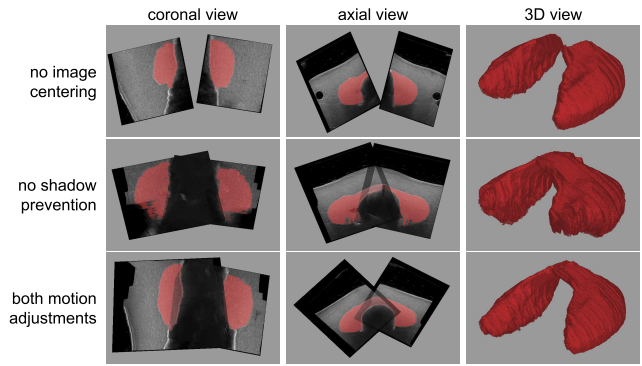}
    \caption{Qualitative results of the thyroid segmentation with shadow correction but without image centering (top), with image centering but without shadow prevention (middle) and with both motion adjustments (bottom).}
    \label{fig:shadow_non_center}
\end{figure}

\subsection{Comparison to conventional 2D US volumetry}

\begin{figure}
   \centering
   \includegraphics[width=\linewidth]{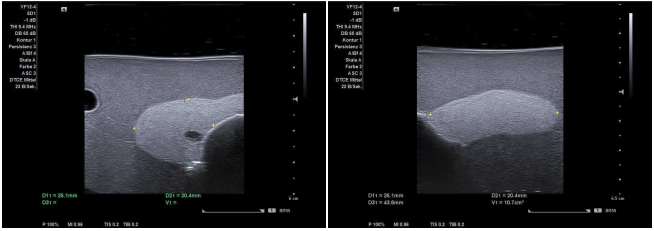}
    \caption{Example of a 2D US measurement of a thyroid lobe. The length of the lobe is measured in all three directions (yellow dotted line).}
    \label{fig:2DUS}
\end{figure}

In the conventional 2D US approach for thyroid volumetry, volunteers were asked to measure the longest line in all three directions of both thyroid lobes in their chosen 2D US images as shown in Figure~\ref{fig:2DUS}. The volume was then computed with an ellipsoid formula 
\begin{align}
    V = c \cdot m_1 \cdot m_2 \cdot m_3 ,
\end{align}
where $c = 0.48$ is the commonly used correction factor and $m_1, m_2$ and $m_3$ are measured lengths in each direction. This calculation is applied per lobe.

In the robotic scan, the volunteers placed the US probe on the phantom in a way that the thyroid lobe was visible and relatively centered. The robot then scanned the lobe as described in subsection~\ref{subsec::Movement}. Each volunteer performed the conventional and the proposed method three times. \\
The results, which are summarized in Table~\ref{table_2D_comparison}, show a significant mean error reduction from $20.85\pm16.10$ to $8.23\pm3.10$ in volumetric estimations between conventional 2D US and our proposed robotic method for the expert group. In both groups, the average volume is closer to the measured ground truth and the standard deviation is smaller, showing closer volumetric results between the scans. Furthermore with our proposed robotic method, the difference in average volume estimations between experts and non-experts is marginal. 
 
\begin{table}[h]
\caption{Comparison between conventional 2D volumetry and our proposed method. Volume is given as mean$\pm$std (ml).}
\label{table_2D_comparison}
\begin{center}
\begin{tabular}{lcc}
\toprule
method & experts & non-experts\\
\midrule
conventional (ml) & 23.75$\pm$4.83 & 18.56$\pm$4.47\\
 \midrule
robotic (ml) & \textbf{32.48}$\pm$\textbf{0.93} & \textbf{31.48}$\pm$\textbf{1.29}\\
\midrule
ground truth (ml) & 30.01 & 30.01\\
\midrule
mean measurement error  &  &  \\
(conventional) (\%) & 20.85$\pm$16.10 & 38.16$\pm$14.89 \\
mean measurement error  &  &  \\
(robotic) (\%) & 8.23$\pm$3.10 & 5.76$\pm$4.30 \\
mean error improvement with  &  & \\
the robotic method (\%) & 61 & 85\\
\bottomrule
\end{tabular}
\end{center}
\end{table}

\section{Discussion}

As mentioned in the state of the art section, other works also presented robotic-based systems for US scanning of the thyroid. Huang et al. \cite{huang} use a robotic arm to acquire high-quality 3D US scans of a thyroid phantom and other phantoms. The US images were then evaluated based on their quality. Kojcev et al. \cite{Kojcev} compared the task of thyroid volumetry based on 2D US and based on a robotic acquisition. The robotic acquisition was executed based on a pre-planned trajectory, and the volume estimation was obtained by manually measuring 2D projections in a 3D volume. In contrast, we offer an automatic approach for the combination of robotic US scanning and thyroid volumetry. 

Tracked 3D US is a different method to acquire 3D US compoundings from an US scan. In such an approach, the US probe is tracked with an optical or electromagnetic tracker, but the scan itself is executed manually. The segmentation part of the proposed framework can therefore be applied to tracked 3D US as well. Our own investigations in this direction have shown good segmentation results on volunteers already \cite{kroenke2021tracked}. However, the tracked US acquisitions are still executed manually, allowing for variability in acquisitions, depending on different operators, human fatigue, and other factors. Furthermore, accuracy measurements show a more precise tracking through robotic acquisitions \cite{STURZ20176863}, compared to optical \cite{doi:10.3109/10929081003647239} or electromagnetic tracking \cite{6810177}.

To show the reliability of the CT ground truth segmentation we analyzed the variability in this segmentation. For that the CT of the phantom was segmented eight times by medical imaging experts. Mean and standard deviation of the volume are $29.54 \pm 0.59 \text{mm}^3$, respectively. The standard deviation is significantly lower than in the US acquisitions (see Table~\ref{table_2D_comparison}). Therefore, this variability can be neglected when comparing to US-based results and it is reasonable to use one CT segmentation as ground truth.

\subsection{Translation to clinical practice}
This manuscript shows an initial work for robotic US-based thyroid volumetry. In future works some aspects have to be considered to translate this work to clinical practice. Currently, the network is overfitted to the phantom to allow for a good acquisition.This overfitting was necessary because the domain shift between US images of the phantom and real thyroids is large. For the transition to real thyroid scans this network should be re-trained, which has successfully shown to be feasible in \cite{kroenke2021tracked}. Patient neck anatomies also show a high variability. Part of the thyroid can, for example, lie below the clavicle or patients can exhibit goiters. These differences could require additional path corrections based on specific neck anatomies in a clinical setup. Furthermore, force analysis in x- and y-direction could be included, as well as a more complex shadow prevention. In case of a goiter, the neck for example shows a convex surface which increases the scanning difficulty. A solution could be to include a US probe monitoring which enforces a normal positioning of the US probe to the surface.

\section{Conclusion}

In this paper, we introduced, to our knowledge, the first full framework for robotic and automatic 3D  thyroid US imaging and volumetry. Our system is capable of scanning a thyroid phantom with a sole interaction by the user when defining the initial position for the US sweep per lobe. Next to the computed volume, a 3D compounded US image is generated which can be used for clinical follow-up.

We further introduce means for compensating shadows at the sides of the US image and keep the organ of interest, here a thyroid lobe, centered in the US tomographic plane. By doing so, we are the first group to offer an online trajectory definition based solely on the US images. An ablation study indicates that both image-guidance methods for the robotic movement improve the volume calculation in our phantom setup.

Finally, in a user study which compares our robotic method with 2D US for volumetry, we prove that our system can outperform volumetry using 2D measurements and the ellipsoid formula both in an expert and non-experts group. Further, when using our proposed approach we show that in our phantom setup non-experts and experts can produce comparable results.

Being a pioneering work to automate 3D thyroid US imaging and volumetry, we acknowledge that the presented evaluation is a first phantom study and the phantom US images differ from real patient data. However, we strongly believe that our robotic-based pipeline can pave the way towards more accurate 3D thyroid US imaging and volumetry in real patients and will inspire automatic US measurements with collaborative robotic support for further anatomies.

The presented advances therefore are a positive step towards automatizing and standardizing thyroid volumetry which can moreover personalize further radioiodine therapy in hyperthyroidism. To be able to translate the system to clinical practice, volunteer studies, as well as patient studies, are planned as next steps in such endeavor.

\addtolength{\textheight}{-12cm}   
                                  
\section*{Acknowledgment}

The authors would like to thank the Nuclear Medicine department at Klinikum rechts der Isar, in particular, Prof. Klemens Scheidhauer, Julian Petzold, Dr. Susan Notohamiprodjo, Markus Kr\"onke and Lisena Cala  for participating as experts in the evaluation.

\bibliographystyle{plain}
\bibliography{references}

\begin{thebibliography}{10}

\bibitem{Brunn}
J~Brunn, U~Block, G~Ruf, I~Bos, W~P Kunze, P~C Scriba, and Janusz Myśliwiec.
\newblock Volumetric analysis of thyroid lobes by real-time ultrasound
  (author's transl).
\newblock {\em Deutsche medizinische Wochenschrift (1946)}, 10 1981.

\bibitem{busam2016quaternionic}
Benjamin Busam, Marco Esposito, Benjamin Frisch, and Nassir Navab.
\newblock Quaternionic upsampling: Hyperspherical techniques for 6 dof pose
  tracking.
\newblock In {\em 2016 Fourth International Conference on 3D Vision (3DV)},
  pages 629--638. IEEE, 2016.

\bibitem{doi:10.3109/10929081003647239}
Robert Elfring, Matías de~la Fuente, and Klaus Radermacher.
\newblock Assessment of optical localizer accuracy for computer aided surgery
  systems.
\newblock {\em Computer Aided Surgery}, 15(1-3):1--12, 2010.
\newblock PMID: 20233129.

\bibitem{esposito2016multimodal}
Marco Esposito, Benjamin Busam, Christoph Hennersperger, Julia Rackerseder,
  Nassir Navab, and Benjamin Frisch.
\newblock Multimodal us--gamma imaging using collaborative robotics for cancer
  staging biopsies.
\newblock {\em International journal of computer assisted radiology and
  surgery}, 11(9):1561--1571, 2016.

\bibitem{6810177}
Alfred~M. Franz, Tamás Haidegger, Wolfgang Birkfellner, Kevin Cleary, Terry~M.
  Peters, and Lena Maier-Hein.
\newblock Electromagnetic tracking in medicine—a review of technology,
  validation, and applications.
\newblock {\em IEEE Transactions on Medical Imaging}, 33(8):1702--1725, 2014.

\bibitem{guha_roy_quicknat_2019}
Abhijit Guha~Roy, Sailesh Conjeti, Nassir Navab, Christian Wachinger, and
  {Alzheimer's Disease Neuroimaging Initiative}.
\newblock {QuickNAT}: {A} fully convolutional network for quick and accurate
  segmentation of neuroanatomy.
\newblock {\em Neuroimage}, 186:713--727, February 2019.

\bibitem{hennersperger2017towards}
Christoph Hennersperger, Bernhard Fuerst, Salvatore Virga, Oliver Zettinig,
  Benjamin Frisch, Thomas Neff, and Nassir Navab.
\newblock Towards mri-based autonomous robotic us acquisitions: a first
  feasibility study.
\newblock {\em IEEE transactions on medical imaging}, 36(2):538--548, 2017.

\bibitem{huang}
Qinghua Huang, Jiulong Lan, and Xuelong Li.
\newblock Robotic arm based automatic ultrasound scanning for three-dimensional
  imaging.
\newblock {\em IEEE Transactions on Industrial Informatics}, 15(2):1173--1182,
  2019.

\bibitem{kaminski}
Jakub~T Kaminski, Khashayar Rafatzand, and Haichong~K Zhang.
\newblock Feasibility of robot-assisted ultrasound imaging with force feedback
  for assessment of thyroid diseases.
\newblock {\em Proceedings of SPIE--the International Society for Optical
  Engineering}, 2020.

\bibitem{kim}
Yeoun~Jae Kim, Jong~Hyun Seo, Hong~Rae Kim, and Kwang~Gi Kim.
\newblock Development of a control algorithm for the ultrasound scanning robot
  (nccusr) using ultrasound image and force feedback.
\newblock {\em The international journal of medical robotics + computer
  assisted surgery : MRCAS}, 13, 2017.

\bibitem{Kojcev}
Risto Kojcev, Ashkan Khakzar, Bernhard Fuerst, Oliver Zettinig, Carole Fahkry,
  Robert DeJong, Jeremy Richmon, Russell Taylor, Edoardo Sinibaldi, and Nassir
  Navab.
\newblock On the reproducibility of expert-operated and robotic ultrasound
  acquisitions.
\newblock {\em International journal of computer assisted radiology and
  surgery}, 06 2017.

\bibitem{kroenke2021tracked}
Markus Kr\"onke, Christine Eilers, Desislava Dimova, Melanie K\"ohler, Gabriel
  Buschner, Lilit Mirzojan, Lemonia Konstantinidou, Marcus~R. Makowski, James
  Nagarajah, Nassir Navab, Wolfgang Weber, and Thomas Wendler.
\newblock Tracked 3d ultrasound and deep neural network-based thyroid
  segmentation reduce interobserver variability in thyroid volumetry, 2021.

\bibitem{STURZ20176863}
Yvonne~R. Stürz, Lukas~M. Affolter, and Roy~S. Smith.
\newblock Parameter identification of the kuka lbr iiwa robot including
  constraints on physical feasibility.
\newblock {\em IFAC-PapersOnLine}, 50(1):6863--6868, 2017.
\newblock 20th IFAC World Congress.

\bibitem{Szumowski}
Piotr Szumowski, Małgorzata Mojsak, Saeid Abdelrazek, Monika Sykała, Anna
  Amelian-Fiłonowicz, Dorota Jurgilewicz, and Janusz Myśliwiec.
\newblock Calculation of therapeutic activity of radioiodine in graves' disease
  by means of marinelli's formula, using technetium ((99m)tc) scintigraphy.
\newblock {\em Endocrine}, 12 2016.

\bibitem{Taylor}
Peter Taylor, Diana Albrecht, Anna Scholz, Gala Gutierrez-Buey, John Lazarus,
  Colin Dayan, and Onyebuchi Okosieme.
\newblock Global epidemiology of hyperthyroidism and hypothyroidism.
\newblock {\em Nature Reviews Endocrinology}, 14, 03 2018.

\end{thebibliography}

\end{document}